# HOW TO MINIMIZE THE ENERGY CONSUMPTION IN MOBILE AD-HOC NETWORKS

Abdellah Idrissi

Laboratory of Computer Science,
Department of Computer Science, Faculty of Sciences
Mohammed V University - Agdal - Rabat – Morocco

Email: idriab@gmail.com

## ABSTRACT

*In this work we are interested in the problem of energy management in Mobile Ad-hoc Network (MANET). The solving and optimization of MANET allow assisting the users to efficiently use their devices in order to minimize the batteries power consumption. In this framework, we propose a modelling of the MANET in form of a Constraint Optimization Problem called COMANET. Then, in the objective to minimize the consumption of batteries power, we present an approach based on an adaptation of the Dijkstra's algorithm to the MANET problem called MANED. Finally, we expose some experimental results showing utility of this approach.*

## 1. INTRODUCTION

The Constraint Network (CN) called also Constraint Satisfaction Problem (CSP) is initially introduced in [1]. It is proven more and more promising to model and solve a large number of real problems. A lot of approaches using constraint reasoning have proposed to solve CN (see for example [2, 3, 4]). A Constraint Network (CN) is defined by the triplet $(X , D, C)$, where $X = \{X1, ..., Xn\}$ is the set of $n$ variables; $D = \{D1, ..., Dn\}$ is the set of $n$ domains of values; $Di$ is the domain of values of the variable $Xi$ and $C = \{C1, ..., Ce\}$ is the set of $e$ constraints of the problem. Solving a CSP consists in assigning values to variables in order to satisfy all the constraints.

Various real problems can be represented in form of a CSP. In this paper, we are interested more particularly in the problem of optimization expressed within the framework of Valued Constraint Satisfaction Problem (VCSP) [2]. We model the Mobile Ad-hoc NETwork problem (MANET) in form of a VCSP and apply VCSP techniques to solve and optimize it. The resolution and optimization of mobile ad-hoc networks permit to assist the users to efficiently use their devices when transmitting messages. This can contribute to minimize the energy consumption of devices since the combinatorial optimization problems make it possible to select the best combination among all those possible.

A mobile device can communicate directly with another device if it is in its range of transmission. Beyond that, the intermediate devices play the part of routers to relay the messages jump by jump. The path between a source device and a destination device can imply several jumps without wire (figure 1).

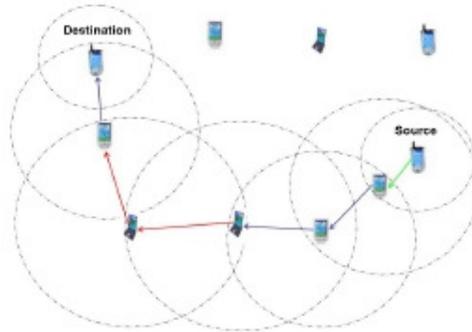

Figure 1: An example of routers relaying messages jump by jump. It shows also a heterogeneous system.

As shown above, wireless ad-hoc networks are limited on resources, such as batteries power. Thus, to efficiently obtain optimal results of its use, the users can employ algorithmic methods in order to minimize the resource consumption.

## 2. RELATED WORK

We present a quick review of the most interesting results in the area of power control for Mobile Ad-hoc NETworks (MANET). Many papers in the area have been concentrated on the development of new protocols that can minimize the consumed power. For example, the authors in [5] give a new protocol for power control, based on information available through lower level network layers. Another approach for power control is presented by Kawadia and Kumar in [6]. Two protocols are proposed, in which the main technique used is the clustering of mobile units according to some of its features.

Three mixed integer programming formulations are presented in Das et al. [7] for the optimization problem of broadcasting a message from a source device to all the other devices with minimum energy. The problem is called MPB (Minimum Power Broadcast) and is shown NP-hard in [8]. So although any standard IP techniques can be used to solve the problem modelled by integer programming, optimal solution can be expected only for problem instances of relatively small size.

Heuristic approaches have to be used to find sub-optimal solutions for hard problems of large size. Wieselthier et al. [9] described a constructive algorithm called BIP (Broadcast Incremental Power). In this algorithm, new devices are added to the tree using a minimum incremental cost heuristic. Marks II et al. [10] presented an evolutionary approach using genetic algorithms together with methods for generating initial solutions. Das et al. proposed an ant colony system approach [11] and a local search heuristic called r-shrink procedure [12] for improving solutions obtained using fast sub-optimal algorithms in wireless networks such as BIP (Broadcast Incremental Power).

The above approaches deal with the cases where the communication between two devices is not necessarily symmetric, i.e. given two devices $i$ and $j$, device $i$ can send a message to device $j$, but device $j$ cannot necessarily send any message to device $i$, since $i$ and $j$ don't necessarily have the same energy. The symmetric case is treated by Montemanni and Gambardella [13] who presented two mixed integer programming formulations for the minimum power symmetric connectivity problem and some valid inequalities for the polytopes associated. The heuristic algorithm is based on the simulated annealing paradigm.

Another minimum energy problem consists in minimizing energy for sending $k$ messages in a MANET, each message from a source device to a destination device. The problem is called PCADHOC (Power Control problem in AD-HOC networks). Carlos and Pardalos proposed in [14] a model for this problem. They proposed a linear integer programming model, which is used to find lower bounds of the amount of required power, and a VNS (Variable Neighborhood Search) local search algorithm and its distributed version to solve the problem.

In this work we are interested in the problem of minimizing energy for sending one message from a source device to a destination device. We treat this problem as a special and basic case of MPB (there is only one destination device) and PCADHOC ($k$=1). We first introduce a constraint model to minimize the amount of power required by network users at a specific time period. The resulting problem is called the Constraint Optimization problem in Mobile Ad-hoc NETwork (COMANET). We assume that a fixed amount of data must be sent from a source device to a destination device and try to determine the optimal amount of power necessary to do this. A new algorithm for this problem is then given.

## 3. A CONSTRAINT OPTIMIZATION MODEL FOR THE MOBILE AD-HOC NETWORK (COMANET)

### 3.1. Problem Formulation

We assume at an instant $t$ a fixed N-device network in which a source device can send a message to a destination device. Any device of the network can be considered as a source or a destination and any device can be used as a relay device to reach other devices in the network. In the rest of this paper, we consider that the communication between two devices is not necessarily symmetric, i.e. given two devices $i$ and $j$, device $i$ can send a message to device $j$, but device $j$ cannot necessarily send any message to device $i$, since $i$ and $j$ don't necessarily have the same energy. We assume that if device $vi$ transmits to device $vj$, then all devices closer to $vi$ than $vj$ also receive the transmission. As described in [12], the minimum transmitter power, $Enij$, that enables device $vi$ to send information to device $vj$ is proportional to $[d(vi, vj)]^{\alpha}$, where $d(vi, vj) = [(xi - xj)^2 + (yi - yj)^2 + (zi - zj)^2]^{1/2}$ is the Euclidean distance between devices $vi$ and $vj$, $(xi, yi, zi)$ (resp. $(xj, yj, zj)$) are the coordinates of device $vi$ (resp. $vj$) and $\alpha$ is a channel-loss exponent that usually lies between 2 and 4, the exact value depends on the nature of the signal-propagation medium. We consider that the proportionality is equal to 1 and therefore $Enij = [d(vi, vj)]^{\alpha}$ ; we assume that $\alpha = 2$.

We consider that $En(l)$ is the transmitter power consumed by a device if it operates at power level $l$, and denote by $N(vi, l)$ the set of wireless units around $vi$ that can be reached by unit $vi$ if it operates at power level $l$. $N(vi, l)$ is also called transmission neighborhood of $vi$. This finite set has $\delta(vi, l) = |N(vi, l)|$ elements and increases as a function of the power level of device $vi$. Our objective is to find the minimum power level necessary to send data from a source device to a destination device. However, at an instant $t$, every unit operates at only one power level, which induces a directed weighted graph $(V, E)$ where $V$ is the set of all units, and $E$ is the set of edges determined by the power level of each unit as follows: edge $(vi, vj) \in E$ if and only if device $vi$ operates at level $l$ and device $vj$ is in $N(vi, l)$. The cost (or weight) of edge $(vi, vj)$ is then $En(l)$.

## 3.2. Problem Modelling

We want to determine the power level of each unit such that the minimum of energy is consumed in the MANET when sending a message from a device $s$ to a device $d$ at an instant $t$. For this purpose, we model the mobile ad-hoc network in form of a constraint optimization COMANET = $(X, D, C)$ where:

- $X = \{vil, xij\}$ ($i, j$=1, ..., $n$; $l$=1,...,$L$). The binary variable $vil$ is defined as $vil = 1$ if and only if unit $vi$ is at the $lth$ power level. The variable $xij$ is a binary variable defined as $xij = 1$ if and only if edge $(vi, vj)$ is in the graph $(V, E)$ induced by the units and their power levels.

- $D = \{0, 1\}$ is the domain of values for binary variables $vil$ and $x_{ij}$.

- $C$ is the set of constraints of the problem. We can formulate them as follows:

$$\sum_{l=1}^{L} v_{il} = 1 \qquad (1)$$

$$x_{ij} d^{\alpha}(i, j) \leq \sum_{l=1}^{L} v_{il} En(l) \qquad (2)$$

$$NbP_{sd} > 0 \qquad (3)$$

where $NbP_{sd}$ is the number of paths existing between devices $s$ and $d$. Generally, we define the number of paths, existing between devices $v_i$ and $v_j$, by the following equation:

$$NbP_{ij} = \sum_{S=is_1s_2...s_mj} \prod x_{is_1} x_{s_1s_2}...x_{s_mj} \qquad (4)$$

Constraint (1) ensures that for each device exactly one power level is selected. Constraint (2) defines the relationship between variables $xij$ and $vil$. It means that the device $vj$ can be reached by device $vi$ if device $vi$ operates at power level $l$ and device $vj$ is in $N(vi, l)$, i.e. if edge $(vi, vj)$ is in $(V, E)$ then $Enij = d^{\alpha}(i, j) < En(l)$. Constraint (3) ensures that, for a source-destination pairs $(vs, vd)$, there is at least a feasible path leading from $vs$ to $vd$. Note that there can be several paths connecting two devices $s$ and $d$. The $k^{th}$ path between two devices $s$ and $d$ (denoted by $P^{th}s$) is a sequence $Sk = <s, S1, S2,, …, Sm, d>$ of distinct devices such that $s = S0$, $d = Sm+1$ and $(Si-1, Si) \in E$, for all $i \in \{1, 2, ..., m + 1\}$. This means that data can be sent through an edge if and only if it really exists in the COMANET (network) induced by the power levels assigned to each device.

We propose here an optimization method able to give the best utilization of different power levels in order to minimize the consumption of energy. The objective of the optimization is:

$$min \quad \sum_{l=1}^{L} \sum_{i=1}^{n} v_{il} En(l) \qquad\qquad (5)$$

subject to constraints (1)-(3).

Now our constraint optimization is well identified. It remains to apply one of search algorithms (Branch and Bound, Backtracking, Dijkstra's algorithm, etc.) to solve the energy management problem in mobile ad-hoc network (MANET). We chose to adapt the Dijkstra's algorithm and call our search method the MANED algorithm.

## 4. AN ALGORITHM FOR ENERGY MANAGEMENT IN MOBILE AD-HOC NETWORK (MANED)

Our approach solving the problem of minimizing the power consumed is an adaptation of the Dijkstra's algorithm to the mobile ad-hoc network. We call this new algorithm MANED. The MANED algorithm is a hybrid algorithm which allows finding the minimum of energy to consume when sending a message from the device $s$ to the device $d$ in the mobile ad-hoc network. This algorithm is based on the Dijkstra's algorithm [15]. Dijkstra's algorithm, when applied to a graph, quickly finds the shortest path from a chosen source to a given destination. The algorithm finds all shortest paths from the source to all destinations. The graph is made of vertices (or nodes), and edges which link vertices together. Edges are directed and have an associated distance, sometimes called the weight or the cost.

The Algorithm 1 partitions devices in two distinct sets, the set of untreated devices and the set of treated devices. Initially all devices are untreated, and the algorithm ends once the destination device $d$ is in the treated set. A device is considered treated, and moved from the untreated set to the treated set, once its shortest distance from the source has been found. We suppose that all devices are known. The algorithm can be described as follows:

**Algorithm 1** searchPaths( )

1: **for** each device $u$ of the set of devices to treat **do**
2:   add $u$ to the set of the devices treated;
3:   **for** each successor $v$ of $u$ **do**
4:     **if** ($v$ not treated yet) **then**
5:       **if** (cost($v$) > cost($u$) + cost($u, v$)) **then**
6:         cost($v$) = cost($u$) + cost($u, v$);
        add $v$ to the set of the devices to treat;
        predecessor = $u$;
7:       **end if**
8:     **end if**
9:   **end for**
10: **end for**

Algorithm 2 first generates the directed weighted graph ($V, E$). If one or several connections are possible between two devices $vi$ and $vj$ , then edge ($vi, vj$ ) is in $E$ and its cost (or weight) is $En(l)$, where $l$ is the lowest power level such that the connection from $vi$ to $vj$ is possible. It then applies algorithm 1 to find the cheapest path from the source device s to the destination device d.

**Algorithm 2** MANED(s, d, m, R)

1: /* m is a message to send from the source device $s$ to the destination device $d$ in the
   network $R$ */
2: **for** each device $u$ **do**
3:   **for** each communication level $l$ supported **do**
4:     create connections with the neighbors of $u$ with which a communication is
       possible, depending on the level $l$;
5:   **end for**
6: **end for**
7: generate the ad-hoc network called here COMANET;
8: specify a source $s$ and a destination $d$;
9: apply the algorithm 1 in order to find the path from $s$ to $d$ which consumes the
   minimum of energy;

## 5. IMPLEMENTATION AND EXPERIMENTAL RESULTS

We suppose that each device has one or more communication levels. Each level has a specific range of transmission. Each device can be localized in the space by its corresponding coordinates (x, y, z). In order to simplify the graphic representation, in this paper without any loss of generality, we assume that each device supports at most 3 levels of communication $l_1$, $l_2$ and $l_3$.

We generate a random ad-hoc network in a space of size (X, Y, Z) with $0 \leq x \leq X$ , $0 \leq y \leq Y$ and $0 \leq z \leq Z$ . Moreover, in order to allow only a 2D representation, we will not consider in this paper the third coordinate $z$ (i.e. $z = 0$). For the purpose of simplifying the research of the visible neighbors, we divide this space into sectors. The size of a sector is a parameter. The ad-hoc network of $N$ devices then will be randomly generated. Each device supports until 3 communication levels as described above. This generation produces quite heterogeneous and random networks.

The ad-hoc network is represented in form of a constraint optimization noted COMANET with each variable (device) $u$ being complete sub-graph constituted by the power levels (figure 2). With each value $l$ is associated a cost with which the device send a message. In addition, a device can change its level to send the message when necessary (for example from $l_2$ to $l_3$ in the device $v_4$ of figure 2). It is also possible to associate a cost $C_b$ to this swing of level (for example $C_b = 1$). We assume that $C_d$ is the cost associated to the destination device and suppose that $C_d = 2$.

We generate N devices distributed in current space in a random way. Each device supports at least one level of communication supporting the weakest range of transmission.

In the example of figure 2, the paths from source device v1 which send information to destination device v5 is marked by arrows. The minimum total cost computed and its corresponding path is also showed.

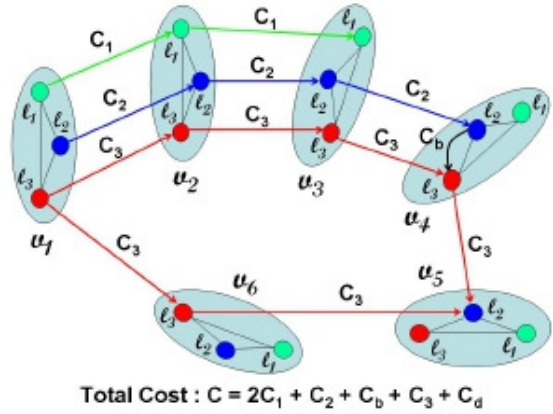

Figure 2. Example of a network constituted by six devices v1, v2, v3, v4, v5 and v6 where each device contains three energy levels l1, l2 and l3.

Concretely, we use probability $pr(li)$ to decide if power level $li$ ($i = 1, 2, 3$) is supported by a device:

$$pr(l1) = 1 \qquad (6)$$

$$pr(l2) = pr(l1) * 3/4 \qquad (7)$$

$$pr(l3) = pr(l2) * 1/2 \qquad (8)$$

The cost of a level is calculated using the following two functions (as in [10]):

$$d(li) = li * (li + 1) \qquad (9)$$

$$C(d) = d^2 \qquad (10)$$

where $d(li)$ is the distance (in sectors) covered by the level $li$ and $C(d)$ is the cost in energy to cover the distance $d$.

Thus, for every level $li$:

$$Cost(li) = C(d(li)) \qquad (11)$$

After a device was created, it is added to the corresponding sector in space. A device $u$ can be connected to another device $v$ if a communication is possible. The path which one wants to find is the path of a device $s$ towards the device $d$ which consumes the less possible energy. The total cost $C$ of such path is thus:

$$C = min [Cost(ls) + \sum (Cost(lr)) + nb * Cb + Cost(ld)] \qquad (12)$$

where $r$ being intermediate devices (i.e. routers); $ls$, $lr$ and $ld$ are operational power levels; $nb$ is the number of swings and $nb * Cb$ is the total cost related to the swings of energy levels which are necessary (see an example in figure 2). The MANED algorithm (i.e. algorithm 2) is then performed to determine the path consuming the minimum of energy.

In the example of figure 2, following the MANED algorithm, the minimum total cost for transmitting information from source device $v1$ to destination device $v5$ is C = 2*$C1$ + $C2$ + $C3$ + $Cb$ + $Cd$ = 2*$2^2$ + $6^2$ + $12^2$ + 1 + 2 = 191 corresponding to the path formed by green colour between $v1$ and $v2$, also green colour between $v2$ and $v3$, blue colour between $v3$ and $v4$, black colour inside the device $v4$ (swing) and finally red colour between $v4$ and $v5$.

A panel makes it possible to draw space; its sectors and the ad-hoc network which this space contains (figure 3). In this figure the number of sectors is $27^2$, the size of one sector is $26^2$ pixels and the total devices generated is only 50 in order to allow a good vision to the reader. Each device is indicated by a feature with one, two or three connectors representing the levels of communication which are supported. Possible connections are mentioned in green, blue or red depending on the range of transmission relating to the level. Green means low level power, blue means middle level power and red means high level power. If a path were required from device $s$ to device $d$, this one will be printed in fat. In figure 3, we assume that there are three source devices $s1$, $s2$ and $s3$ which send message respectively to three destinations devices $d1$, $d2$ and $d3$. The calculation of the minimum power necessary to send information respectively from source devices $s1$=45[10;521;17], $s2$=38[678;162;0] and $s3$=13[57;408;0] to destination devices $d1$=49[682;16;0], $d2$=35[135;678;0] and $d3$=42[341;116;0] of the network is represented by the path which is printed in fat with the colour corresponding to the operational power level.

In figure 3, we assume for example that source device $s1$ is device 45[10; 521; 0] (where 10, 521 and 0 are respectively coordinates x, y and z of the device 45) and the destination device $d1$ is 49[682; 16; 0]. The path consuming the minimum energy when sending data from source device $s1$ = 45[10; 521; 0] to destination device $d1$ = 49[682; 16; 0] is as follows: 45[10; 521; 17] $-C2->$ 12[107; 501; 0] $-C2->$ 37[261; 433; 0] $-Cb->$ 37[261; 433; 0] $-C1->$ 30[309; 456; 0] $-C1->$ 14[362; 398; 0] $-C1->$ 28[366; 358; 0] $-Cb->$ 28[366; 358; 0] $-C3->$ 43[624; 76; 0] $-Cb->$ 43[624; 76; 0] $-C1->$ 49[682; 16; 0] $-Cd->$ 49[682; 16; 0].

Total Cost: $C$ = (4 * $C1$) + (2 * $C2$) + (1 * $C3$) + (3 * $Cb$) + $Cd$ = (4 * $2^2$) + (2 * $6^2$) + (1 * $12^2$) + (3 * $1$) + 2 = 237 (figure 3). Note that between devices 37[261; 433; 0], 30[309; 456; 0], 14[362; 398; 0] and 28[366; 358; 0], on the one hand and between devices 43[624; 76; 0] and 49[682; 16; 0] on the other hand, the colour used is green.

In the same way the colour used between devices 45[10; 521; 17], 12[107; 501; 0] and 37[261; 433; 0] is blue and finally the colour used between devices 28[366; 358; 0] and 43[624; 76; 0] is red. It means that alternately we used, following cases, either low level power (green colour corresponding to power cost $2^2$ = 4), middle level power (blue colour corresponding to power cost $6^2$ = 36) or high level power (red colour corresponding to power cost $12^2$ = 144) in order to consume in the total the less possible power energy for sending information from $s1$ to $d1$. In the path mentioned above the devices 37[261; 433; 0], 28[366; 358; 0] and 43[624; 76; 0] are repeated twice, it means that there is a swing of level at those devices. We don't count the power for the first occurrence of those devices but we only add a swing cost $Cb$ ($Cb$ = 1 for example) to the total power of the path.

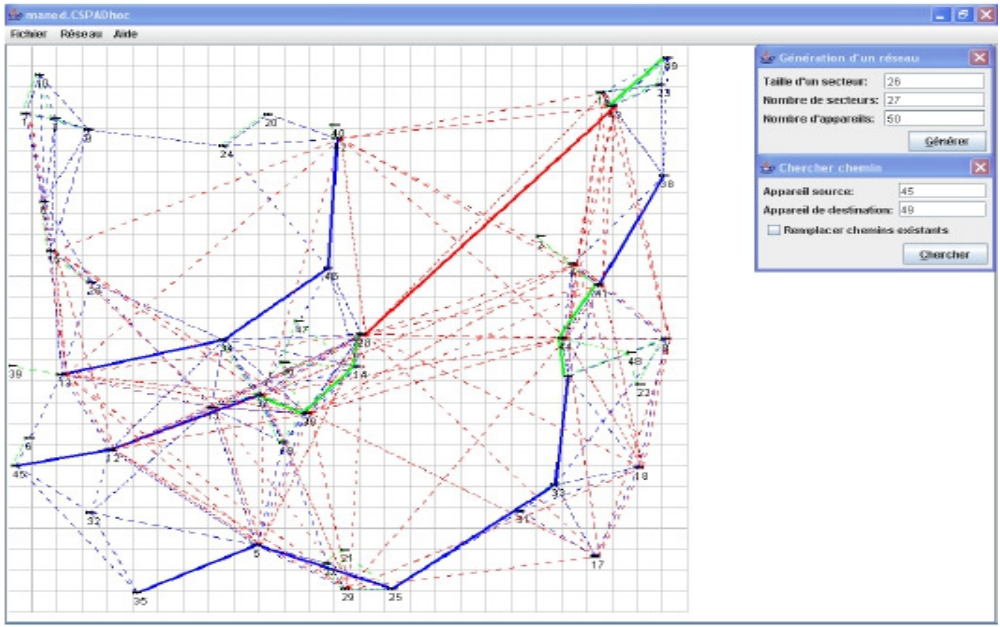

Figure 3. Example of a space, its sectors and the mobile ad-hoc network which this space contains.

In the same figure, there are only the green and the blue colours in the path required when sending data from source device $s2$ = 38[678; 162; 0] to destination device $d2$ = 35[135; 678; 0]. It means that it doesn't need to use the high level power which is represented by the red colour. The path required is: 38[678; 162; 0] $-C2->$ 41[611; 297; 0] $-Cb->$ 41[611; 297; 0] $-C1->$ 44[573; 363; 0] $-C1->$ 7[579; 410; 0] $-Cb->$ 7[579; 410; 0] $-C2->$ 33[566; 544; 0] $-C2->$ 25[398; 673; 0] $-C2->$ 5[258; 619; 0] $-C2->$ 35[135;678;0] $-Cd->$ 35[135;678;0].
Total Cost: $C = 2 * C1 + 5 * C2 + 0 * C3 + 2 * Cb + Cd = 2 * 2^2 + 5 * 6^2 + 0 * 12^2 + 2 * 1 + 2 = 192$ (figure 3).

In the same way, there is only the blue colour in the path required when sending data from source device $s3$ = 13[57; 408; 0] to destination device $d3$ = 42[341; 116; 0]. It means that it doesn't need to use the high level power which is represented by the red colour and also means that the low level power (green colour) isn't enough to send information from device 13[57; 408; 0] to device 42[341; 116; 0]. The path required is: 13[57; 408; 0] $-C2->$ 34[224; 365; 0] $-C2->$ 46[332; 276; 0] $-C2->$ 42[341; 116; 0] $-Cd->$ 42[341; 116; 0].
Total Cost: $C = 0 * C1 + 3 * C2 + 0 * C3 + 0 * Cb + Cd = 0 * 2^2 + 5 * 6^2 + 0 * 12^2 + 0 * 1 + 2 = 110$ (figure 3).

In our experimentation, we generate different networks with 100, 500, 1000, 2000, 3000, 4000 and 5000 devices. The generation of the space, sectors and networks with these numbers of devices added to the operation of search of the path required is generally found in less than one second. However, if we don't count the time required for generating space, sectors and networks, then the necessary time to seek only the path consuming minimum energy power between the source device and the destination device does generally not exceed a few ms for all the sizes of the network.

## 6. CONCLUSION

In this article, we presented a modelling of the mobile Ad-hoc networks problem in form of a constraint optimization problem called COMANET. Thereafter, we proposed an optimization method under constraints for minimizing the power batteries consumption when sending messages from a source device to a destination device. The method presented is based on the adaptation of the Dijkstra's algorithm. The resulting algorithm is called MANED. We presented different experimentations illustrating our approach which can assist users to control and regulate batteries capacities in order to minimize the consumption. The experimental results show that our approach gives very promising results.